\documentclass{article}

\usepackage{xspace}

\PassOptionsToPackage{numbers,sort&compress}{natbib}  
\usepackage[breaklinks,colorlinks,citecolor=green,urlcolor=blue,bookmarks=false]{hyperref}
\usepackage[final]{neurips_2023}

\usepackage[utf8]{inputenc} 
\usepackage[T1]{fontenc}    
\usepackage{hyperref}       
\usepackage{url}            
\usepackage{booktabs}       
\usepackage{amsfonts}       
\usepackage{nicefrac}       
\usepackage{microtype}      
\usepackage{xcolor}         
\usepackage{amsmath}
\usepackage{graphicx}
\usepackage{mathtools}
\usepackage{caption}
\usepackage{subcaption}
\usepackage{makecell}
\usepackage{pifont}
\usepackage{multirow}
\usepackage{algpseudocode}
\usepackage{algorithm}

\usepackage[capitalise,noabbrev,nameinlink]{cleveref}
\crefformat{equation}{#2Equation~#1#3}

\definecolor{CheckGreen}{rgb}{0, 0.55, 0}

\title{Disentangled Interaction Representation for One-Stage Human-Object Interaction Detection}

\author{
  Xubin Zhong$^1$ \quad Changxing Ding$^{1,2}$\thanks{Corresponding author} \quad Yupeng Hu$^1$  \quad Dacheng Tao$^3$\\
  South China University of Technology, China\\
  Pazhou Lab, China\\
The University of Sydney, Australia \\
  \small \texttt{eexubin@mail.scut.edu.cn, chxding@scut.edu.cn, dacheng.tao@sydney.edu.au} \\
}

\begin{document}

\maketitle

\begin{abstract}
Human-Object Interaction (HOI) detection is a core task for human-centric image understanding. Recent one-stage methods adopt a transformer decoder to collect image-wide cues that are useful for interaction prediction; however, the interaction representations obtained using this method are entangled and lack interpretability. In contrast, traditional two-stage methods benefit significantly from their ability to compose interaction features in a disentangled and explainable manner. In this paper, we improve the performance of one-stage methods by enabling them to extract disentangled interaction representations. First, we propose Shunted Cross-Attention (SCA) to extract human appearance, object appearance, and global context features using different cross-attention heads. This is achieved by imposing different masks on the cross-attention maps produced by the different heads. Second, we introduce the Interaction-aware Pose Estimation (IPE) task to learn interaction-relevant human pose features using a disentangled decoder. This is achieved with a novel attention module that accurately captures the human keypoints relevant to the current interaction category. Finally, our approach fuses the appearance feature and pose feature via element-wise addition to form the interaction representation. Experimental results show that our approach can be readily applied to existing one-stage HOI detectors. Moreover, we achieve state-of-the-art performance on two benchmarks: HICO-DET and V-COCO. 
\end{abstract}

\section{Introduction}
\label{sec:1}

Human-Object Interaction (HOI) detection \cite{zhang2021mining,zhong2022towards,lin2022hl} aims to localize interactive human-object pairs in an image and identify the interaction categories between each human-object pair. The obtained results can be formulated  as a set of HOI triplets $<$human, interaction, object$>$. HOI detection has numerous applications in robotics \cite{shan2020understanding,karamcheti2023language,wang2023mimicplay}, human-computer interaction \cite{lv2022deep,balcombe2022human}, and has become an active area of research in computer vision in recent years. However, this task is also very challenging, primarily due to its composite nature, since it requires simultaneous determination of the human and object instance locations, object classes, and categories of interaction. Another challenge pertains to the learning of interaction representations, mainly because the interaction area is usually difficult to define \cite{zhong2021glance, kim2020uniondet}.

\begin{figure}[t]
\centering
\includegraphics[width=1.0\linewidth]{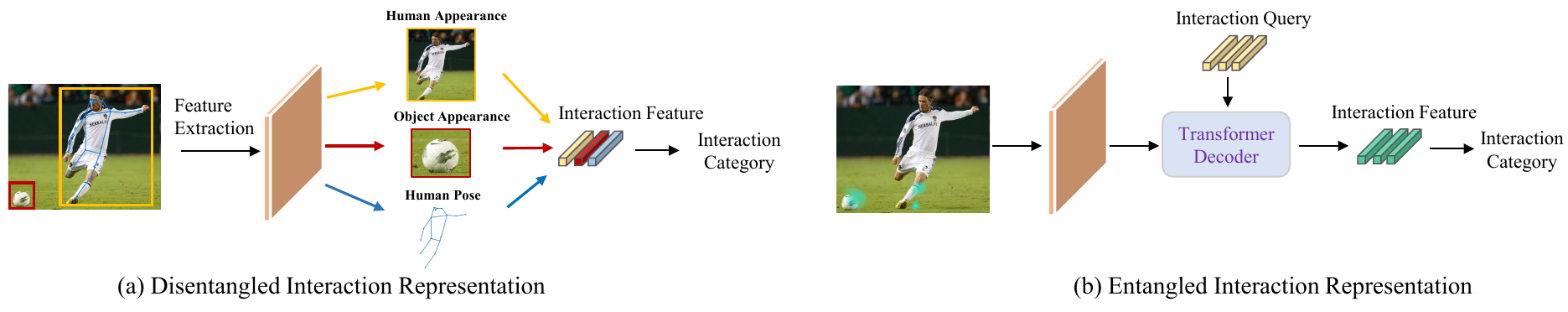}
\caption{Comparisons of interaction representation learning strategies. (a) Conventional two-stage methods compose the interaction representation in a disentangled way, which enables them to explore various interpretable cues for interaction classification. (b) The recent one-stage methods adopt the cross-attention operation to extract image-wide cues useful for interaction classification, resulting in an entangled interaction representation.}
\vspace{-5mm}
\label{Figure:1}
\end{figure}

Conventional two-stage HOI detection methods  \cite{Gao2018iCAN,wan2019pose,zhong2020polysemy} compose the interaction representation in a disentangled manner, which is flexible to include interpretable cues for interaction classification. 
As shown in Fig. \ref{Figure:1}(a),  these approaches usually integrate the human appearance feature, object appearance feature, and human pose feature. The downside of the two-stage methods lies in their low efficiency. In contrast, the recent popular one-stage methods extract the interaction representation efficiently 
\cite{zhang2021mining,tamura2021qpic,zhong2022towards}.  As shown in Fig. \ref{Figure:1}(b), these approaches are usually based on the DEtection TRansformer (DETR) \cite{vaswani2017attention} and adopt the cross-attention operation to search for image-wide cues for interaction prediction. However, the obtained interaction representations are entangled and lack interpretability, which affects their ability to predict interactions. \textbf{Consequently, it is natural to ask whether it is possible for the one-stage methods to benefit from disentangled interaction representations.}
 

Accordingly, we herein propose a novel method that enables one-stage HOI detection methods to extract interaction representations in a disentangled manner, namely, Disentangled Interaction Representation (DIR).  As illustrated in Fig. \ref{Figure:overview}, our method mimics the representation learning strategy of the two-stage methods. First, we propose a novel Shunted Cross-Attention (SCA) training mechanism to explicitly disentangle and extract the human and object appearance features. Specifically, SCA shunts the cross-attention heads in each transformer decoder layer into three groups. It employs different attention-map masks in order to force the three groups of heads to attend to the human appearance, object appearance, and global context features, respectively.  In the inference stage, SCA is removed and therefore introduces no additional computational cost. Second, we introduce a new Interaction-aware Pose Estimation (IPE) task to efficiently extract the human pose feature. IPE shares both the backbone and queries with the interaction prediction task. As the correlations between each body keypoint and different interaction categories vary significantly \cite{li2020pastanet}, we further devise an attention module to capture the importance of each keypoint to the human–object pair of interest. This attention mechanism drives IPE to extract only the specific pose feature that is relevant for interaction classification. Finally, DIR produces the interaction representation by merging the appearance feature and pose feature through element-wise addition.

Our work makes three main contributions. (1) To the best of our knowledge, DIR is the first approach that enables the one-stage HOI detection models to extract disentangled interaction representations. (2) DIR is computationally efficient and plug-and-play; it can be readily applied to existing one-stage HOI detectors. (3) We consistently achieve state-of-the-art performance on two HOI detection benchmarks: HICO-DET and V-COCO.

\section{Related Work}
According to whether the same model is used for both object detection and interaction prediction, existing approaches can be categorized into two-stage and one-stage methods. 

\noindent \textbf{Two-Stage HOI Detection Methods.}
Traditional HOI detection methods \cite{Gao2018iCAN,li2019transferable,wan2019pose,zhong2020polysemy} generally adopt a two-stage framework, which first utilizes an off-the-shelf object detection model \cite{ren2015faster} to detect both human and objects, and then predicts the interaction categories between each human-object pair. The primary advantage of two-stage methods is their flexibility in composing the interaction representation. Multi-stream features can be employed, e.g., the human appearance, object appearance, and human-object spatial configuration \cite{Gao2018iCAN,li2019transferable,wan2019pose}. Moreover, diverse model structures are  devised to enhance interaction representation learning.
For example, Wang et al. \cite{wang2019deep} proposed a spatial attention mechanism to encode contextual features for interaction prediction; while Zhong et al. \cite{zhong2021polysemy} devised a channel attention module to learn polysemy-aware feature. Other works utilize graph convolutional networks to aggregate human and object appearance features \cite{ulutan2020vsgnet}, \cite{gao2020drg}, \cite{xu2019learning}. Moreover, human pose provides fine-grained cues for interaction prediction. Existing two-stage methods usually encode human pose features by utilizing the detected 2D human keypoints \cite{li2019transferable,gupta2019no,zhong2020polysemy}. Some works use this 2D keypoint information to extract a fine-grained human body part feature \cite{wan2019pose,zhou2019relation}; while Li et al.  \cite{li2020detailed} further proposed to use 3D human pose cues to eliminate the ambiguity inherent in 2D HOI detection.  However, the human pose feature is rarely considered by one-stage methods, as it is challenging to directly integrate pose estimation tools into such methods. To the best of our knowledge, IPE is the first approach to encode human pose features for one-stage models.

\noindent \textbf{One-Stage HOI Detection Methods.} 
Due to their advantages in efficiency, one-stage HOI detection methods have become popular in recent years \cite{liao2020ppdm,zhong2021glance,zhang2021mining,chen2021reformulating,tamura2021qpic}. These approaches typically formulate HOI detection as two parallel subtasks in the same model, i.e., human-object pair detection and interaction prediction. Current one-stage methods can be further categorized into interaction point-based methods \cite{liao2020ppdm,zhong2021glance} and DETR-based methods \cite{zhang2021mining,chen2021reformulating,tamura2021qpic}. Both approaches generally employ a single stream structure to learn interaction representation. 
For example, interaction point-based methods tend to extract the interaction feature from the feature backbone using the interaction points \cite{liao2020ppdm}. However, these interaction points are vague and contain very limited information, affecting the performance of these methods in interaction prediction in complex scenes. In contrast, DETR-based methods typically adopt a set of interaction queries, each of which employs the cross-attention mechanism to integrate image-wide information for interaction prediction \cite{zhang2021mining,tamura2021qpic}. 

According to the property of interaction queries  employed, DETR-based one-stage methods can be classified into
(i) approaches that adopt weight-fixed interaction queries during inference \cite{tamura2021qpic,chen2021reformulating,zhong2022towards} and 
(ii) approaches that employ dynamic interaction queries \cite{zhang2021mining,liao2022gen,wu2022mining,kim2022mstr}.
The first class of methods are vulnerable to the weight-fixed  and semantically unclear interaction queries. To address this issue, Iftekhar  et al. \cite{iftekhar2022look}  augmented the interaction queries with additional semantic and spatial cues; while Zhong et. al \cite{zhong2022towards} improved  the model's robustness by training DETR-based methods using hard-positive queries. The second class of methods adopt dynamic and semantically more clear interaction queries, which are usually obtained according to the human and object decoder embeddings. They therefore typically focus on further improving the interaction representation power. For example, Kim  et al. \cite {kim2022mstr}  extracted  multi-scale interaction features using Deformable DETR \cite{zhudeformable}; while Liao et al. \cite{liao2022gen} transferred  knowledge of the CLIP model \cite{clip} to enhance interaction understanding. 
However, as shown in Fig. \ref{Figure:1}, the one-stage methods rarely consider disentangled interaction–representation learning, which may result in the loss of important cues for interaction prediction. \\
\indent In this paper, we propose a novel method that enables one-stage methods to learn disentangled interaction representation. Specifically, we adopt SCA to efficiently extract human and object appearance features, and introduce IPE to efficiently encode interaction-aware human pose feature. In the experimentation section, we further demonstrate that our approach can be easily applied to existing one-stage HOI detectors.

\section{Method}
\label{Method}

\begin{figure*}[t]
	\centering
	\includegraphics[width=1.0\textwidth]{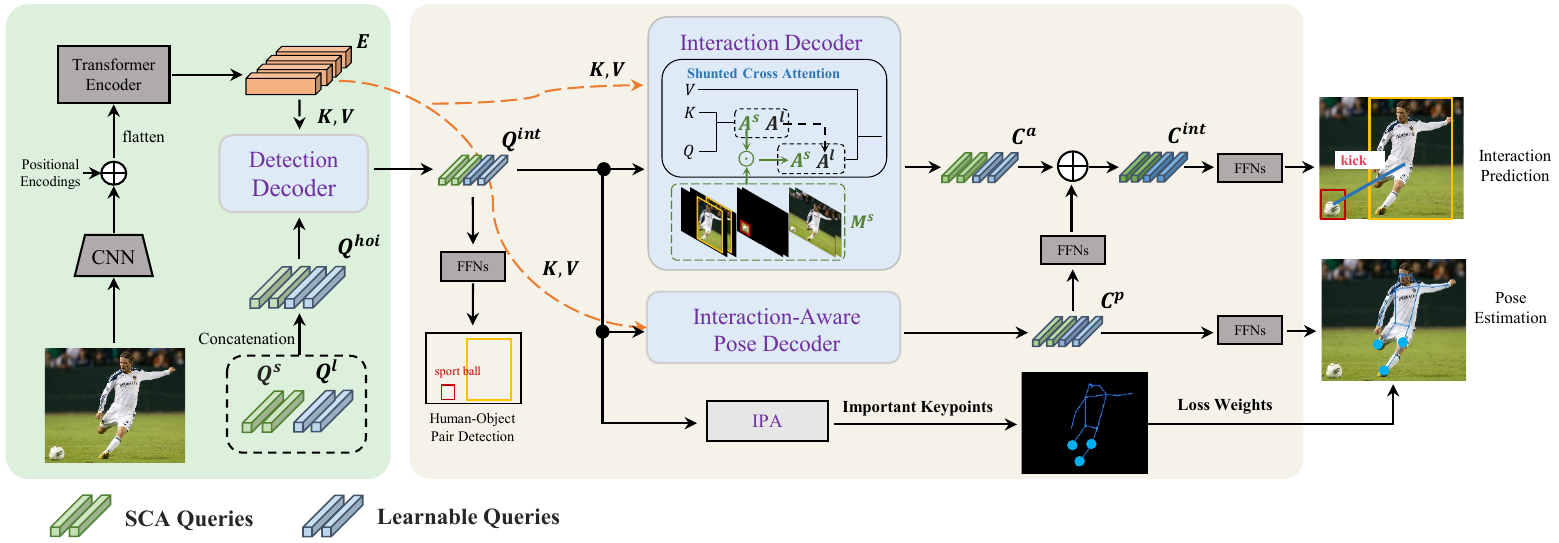}
	\caption{Overview of DIR in the training stage. $\textbf{Q}^{hoi}$ are the queries for the detection decoder, which are responsible for interactive human-object pair detection. $\textbf{Q}^{hoi}$ include both $\textbf{Q}^{l}$ and $\textbf{Q}^{s}$. $\textbf{Q}^{int}$ are the queries for both the interaction and pose decoders, which output the appearance feature $\textbf{C}^{a}$ and the pose feature $\textbf{C}^{p}$, respectively. The final interaction feature is obtained by fusing $\textbf{C}^{p}$ and $\textbf{C}^{a}$ via element-wise addition. $\textbf{A}^{s}$ is modified by shunted head masks $\textbf{M}^{s}$ while $\textbf{A}^{l}$ is not. Both $\textbf{Q}^{s}$ and $\textbf{M}^{s}$ are generated using GT bounding boxes. They are used to regularize the  cross-attention heads to learn human appearance, object appearance, and global feature, respectively.    $\odot$ and $\oplus$ denote the element-wise multiplication and addition operations,  respectively. During inference, $\textbf{Q}^{s}$ and $\textbf{M}^{s}$ are abandoned; only $\textbf{Q}^{l}$ is left for HOI detection (see the supplementary file). Best viewed in color. }
	\label{Figure:overview}
	\vspace{-3mm}
\end{figure*}

DIR is plug-and-play and can be applied to existing DETR-based one-stage HOI detection models \cite{zhang2021mining,liao2022gen}. In this section, we take the representative work CDN \cite{zhang2021mining} as an example. The overall framework of CDN  \cite{zhang2021mining} equipped with DIR for the training phase  is illustrated in Fig.  \ref{Figure:overview}. We also present the framework of DIR during inference in the supplementary file. In the following, we first provide a brief review of our method, which is followed by descriptions of two novel modules, e.g., SCA (Section \ref{sec:SCA}) and IPE (Section \ref{IPE}). Finally, we present the training loss functions for DIR.

\subsection{Overview}
As shown in Fig. \ref{Figure:overview}, DIR comprises a CNN-based backbone, a  transformer encoder, a transformer-based detection decoder, an interaction decoder, a pose decoder, and feed-forward networks (FFNs). Given an image \textbf{I} $\in \mathbb{R}^{H_0 \times W_0 \times 3}$, the CNN backbone and transformer encoder are sequentially employed to extract flattened visual features $\textbf{E} \in \mathbb{R}^{(H \times W) \times D}$. $H \times W$ and $D$ respectively denote the number of the image patches and feature dimension for each patch. DIR then produces the detection embeddings $\textbf{Q}^{int}$ by performing cross-attention between the HOI queries $\textbf{Q}^{hoi}\in \mathbb{R}^{N_{q}\times D}$ and $\textbf{E}$ in the detection decoder.  There are  two types of HOI queries $\textbf{Q}^{hoi}$, i.e. the learnable queries $\textbf{Q}^{l}$ and the SCA queries $\textbf{Q}^{s}$. $N_{q}$ is the total number of $\textbf{Q}^{l}$ and $\textbf{Q}^{s}$. $\textbf{Q}^{int}$ is sent to the FFNs to detect interactive human-object pairs. Besides,  $\textbf{Q}^{int}$ is also utilized as the queries for both the interaction decoder and the pose decoder. In the interests of clarity, we present only the cross-attention operation in one decoder layer. The output embedding $\textbf{C}_{i} \in \mathbb{R}^{N_{q} \times D }$ of the $i$-th decoder layer can be formulated as below:
\begin{eqnarray}
\textbf{C}_{i} = \operatorname{Concat}([\textbf{A}_{j} \textbf{E} \textbf{W}_{j}^V]^T_{j=1}),
\label{attention map1}
\end{eqnarray}
\begin{eqnarray}
\textbf{A}_{j} = \operatorname{Softmax}(Att_{j}(\textbf{Q}, \textbf{C}_{i-1},\textbf{E})),
\label{attention map2}
\end{eqnarray}
where $T$ is the number of cross-attention heads. $\textbf{A}_{j} \in \mathbb{R}^{N_{q} \times (H\times W)} $ is the normalized cross-attention map for the $j$-th head. $\textbf{W}_{j}^V$ is a matrix used for linear projection, and $Att_{j}(\cdot)$ is a function for similarity computation.

\noindent \textbf{Disentangled  Interaction Representation.} As analyzed in Section \ref{sec:1}, recent one-stage HOI detection methods rely solely on a transformer decoder to capture image-wide cues, resulting in an entangled interaction representation. To acquire disentangled interaction representation for one-stage methods,  we propose the Shunted Cross Attention (SCA) and Interaction-aware Pose Estimation (IPE) methods.  
First, SCA is utilized to explicitly integrate the human and object appearance features. To reduce computational cost during inference, we implement SCA in the form of multi-task learning. As shown in Fig.\ref{Figure:overview}, $\textbf{Q}^{s}$ and shunted head masks $\textbf{M}^{s}$ are constructed to perform SCA. During inference, SCA is abandoned and therefore introduces no additional computational cost. Second, IPE is constructed to explore the human pose feature. $\textbf{Q}^{int}$  is used as pose queries for the pose decoder. We  further devise an Interaction-aware Pose Attention (IPA) module to capture important keypoints during training.  Finally, DIR produces the disentangled interaction representation $ \textbf{C}^{int}$  by fusing the appearance feature $ \textbf{C}^{a}$ and pose feature $ \textbf{C}^{p}$:
\begin{eqnarray}
\textbf{C}^{int} = \textbf{C}^{a} \oplus \operatorname{FFN}(\textbf{C}^{p}),
\label{eq:fuse}
\end{eqnarray}
where $\textbf{C}^{int}$ is fed into the interaction classifier to predict interaction scores.

\subsection{Shunted Cross Attention}
\label{sec:SCA}
Human and object appearance features are key components of the interaction representation in traditional two-stage HOI detection methods \cite{hou2020visual, li2020detailed, zhong2020polysemy}.
However, they are rarely explicitly considered in the more popular one-stage methods. There are two main reasons for this. First, the human and object appearance are usually extracted through ROI pooling operations \cite{ren2015faster} in two-stage methods  \cite{hou2020visual, li2020detailed, zhong2020polysemy}, which is a time-consuming task that significantly hampers the inference efficiency of one-stage methods.
Second,  the one-stage approaches train the human-object pair detection and interaction prediction tasks together in each iteration. However, due to the poor-quality of the human/object bounding boxes in the early training epochs, the extracted human and object appearance features are vague,  which impedes the learning of the interaction representation. 

To address the above problem, we propose the SCA approach to disentangle and learn human and object appearance features by means of multi-task learning.  During training, as shown in Fig. \ref{Figure:overview},   SCA employs $\textbf{Q}^{s}$ as the query and shunted attention-map masks $\textbf{M}^{s}$ to extract the above features in the interaction decoder. During inference, $\textbf{Q}^{s}$ and $\textbf{M}^{s}$ are removed and only $\textbf{Q}^{l}$ is used for interaction prediction.

\noindent \textbf{Shunted Head.}  
As shown in Fig. \ref{Figure:overview}, the heads in the cross-attention operation of each interaction decoder layer are shunted into three groups,  which  can be formulated as follows:
\begin{equation}
\textbf{C}_{i} = \operatorname{Concat}([\textbf{A}^{s}_{h} \textbf{E} \textbf{W}_{h}^V]^{T_{h}}_{h=1},
[\textbf{A}^{s}_{o} \textbf{E} \textbf{W}_{o}^V]^{T_{o}}_{o=1}, [\textbf{A}^{s}_{g} \textbf{E} \textbf{W}_{g}^V]^{T_{g}}_{g=1}),
\label{SCA}
\end{equation}
\begin{eqnarray}
\textbf{A}^{s}_{h} = \operatorname{Softmax}(Att_{h}(\textbf{Q}^{s}, \textbf{C}_{i-1},\textbf{E}))\odot M^{s}_h,
\label{HeadH}
\end{eqnarray}
\begin{eqnarray}
\textbf{A}^{s}_{o} = \operatorname{Softmax}(Att_{o}(\textbf{Q}^{s}, \textbf{C}_{i-1},\textbf{E}))\odot M^{s}_o,
\label{HeadO}
\end{eqnarray}
where ${A}^{s}_{h}$($T_h$), ${A}^{s}_{o}$($T_o$), and ${A}^{s}_{g}$($T_g$) stand for the cross-attention maps (the number of cross-attention heads) for the human appearance, object appearance, and global features, respectively. The sum of $T_h$, $T_o$, and $T_g$ is equal to $T$. 
$M^{s}_h$ and $M^{s}_o$ are masks imposed on the attention maps to acquire the human and object appearance features; they are only applied to $\textbf{Q}^{s}$ during training.
With SCA, the multi-head cross-attention in the interaction decoder layers are disentangled to extract the human appearance, object appearance, and global features, respectively.

\noindent \textbf{Ground-Truth Bounding Box.} As discussed above, applying SCA directly to $\textbf{Q}^{l}$ may result in interference with interaction representation learning due to the poor-quality of the predicted bounding boxes. In order to extract high-quality appearance features, we generate $M^{s}_h$ and $M^{s}_o$ according to the Ground-Truth (GT) bounding boxes of human-object pairs. Accordingly, $\textbf{Q}^{s}$ are also generated using coordinates of the GT bounding boxes following \cite{qu2022distillation}. The other design details of SCA are provided in the supplementary file.  

\subsection{Interaction-Aware Pose Estimation}
\label{IPE}
Human-object interaction is strongly correlated with human pose \cite{wan2019pose}. 
Therefore, existing two-stage methods usually employ the human pose feature as one part of the interaction representation \cite{li2019transferable,wan2019pose,zhong2020polysemy}. 
The pose feature is encoded using the body keypoints extracted by off-the-shelf pose estimation models \cite{vitpose2022}. However, one-stage methods are end-to-end and are required to be free from outside tools. Therefore, existing one-stage models necessarily ignore the pose feature. To handle this problem, we propose the IPE method which efficiently extracts the interaction-relevant pose feature.

\noindent \textbf{Human Pose Estimation.}
As shown in Fig. \ref{Figure:overview}, interaction queries $\textbf{Q}^{int}$ are also utilized as the queries for the pose decoder. The produced decoder embeddings are denoted as $\textbf{C}^{p}$:
\begin{eqnarray}
\textbf{C}^{p} = \operatorname{Decoder}(\textbf{Q}^{int},\textbf{E}).
\label{PoseDec}
\end{eqnarray}
$\textbf{C}^{p}$ are then fed to the FFNs to predict body keypoints:
\begin{eqnarray}
\textbf{P} = \operatorname{Sigmoid}{(\operatorname{FFN}}(\textbf{C}^{p})),
    \label{eq:ps}
\end{eqnarray}
where $ \textbf{P}_{i} \in \textbf{P} $, and $ \textbf{P}_{i} \in \mathbb{R}^{K\times 2}$ denotes the coordinates of $K$ keypoints for the $i$-th person.

\noindent \textbf{Interaction-Aware Pose Attention.}
As outlined in previous work \cite{li2020pastanet}, each interaction is potentially related to some important keypoints. To capture important keypoints, we devise a novel attention module, named Interaction-aware Pose Attention (IPA).
As shown in Fig.  \ref{Figure:overview}, the interaction queries  $\textbf{Q}^{int}$ are fed to IPA; IPA then outputs the weights of  keypoints, which can be represented as follows:
\begin{eqnarray}
\textbf{M}^{p} = \operatorname{Softmax}(\operatorname{IPA}(\textbf{Q}^{int})),
\label{eq:IPA}
\end{eqnarray}
where $ \textbf{M}^{p}_{i} \in \textbf{M}^{p}$, and $\textbf{M}^{p}_{i}  \in \mathbb{R}^{K}$ represents the importance scores of the $K$ keypoints for the $i$-th person. $\textbf{M}^{p}$ is only used as pose loss weights for the predicted keypoints during training (in Eq. (\ref{eq:pose_loss})). IPA is constructed by  two successive fully-connected layers, the dimensions of which are both $D$. The first fully-connected layer is followed by a ReLU layer, while the second one is followed by a softmax function. During inference, IPA is removed and therefore introduces no additional cost.

\noindent \textbf{Pose Loss Function.}
We adopt the  commonly-used L1 loss for pose estimation, as follows:
\begin{eqnarray}
\mathcal{L}_{p} =  \frac{1}{N_q} \sum_{i}^{N_q} \| (\textbf{P}_{i} - \textbf{P}^{*}_{i}) \odot \textbf{M}^{p}_{i} \|,
\label{eq:pose_loss}
\end{eqnarray}
where $\textbf{P}_{i}$ and $\textbf{P}_{i}^{*}$ denote the human keypoints predicted  by IPE and one off-the-shelf pose estimation model \cite{vitpose2022}, respectively.  The latter is utilized as the pseudo-label of body keypoints during training.

\noindent \textbf{Overall Loss Function.} We adopt the same loss functions for object detection and interaction prediction as those in CDN \cite{zhang2021mining}. The overall loss function in the training phase can be represented as follows:
\begin{eqnarray}
\centering
\begin{aligned}
 \mathcal{L} =  \alpha \mathcal{L} _{l} + \beta \mathcal{L} _{s} + \gamma  \mathcal{L} _{p},
\end{aligned}
\label{eq:all_loss}
\end{eqnarray}
where
\begin{eqnarray}
\centering
\begin{aligned}
  \mathcal{L} _{l} = \lambda_{b}  \mathcal{L} _{l_{b}}  +  \lambda_{u}  \mathcal{L} _{l_{u}}  +
  \lambda_{c}  \mathcal{L} _{l_{c}} +   \lambda_{a}  \mathcal{L} _{l_{a}},
\end{aligned}
\label{eq:all_loss2}
\end{eqnarray}
\begin{eqnarray}
\centering
\begin{aligned}
  \mathcal{L} _{s} = \lambda_{b}  \mathcal{L} _{s_{b}}  +  \lambda_{u}  \mathcal{L} _{s_{u}}  +
  \lambda_{c}  \mathcal{L} _{s_{c}} +   \lambda_{a}  \mathcal{L} _{s_{a}}.
\end{aligned}
\label{eq:all_loss3}
\end{eqnarray}
$  \mathcal{L}_{l}$ and $  \mathcal{L}_{s}$ denote the loss functions imposed on the predictions of $\textbf{Q}^{l}$  and $\textbf{Q}^{s}$, respectively.
$ \mathcal{L}_{k_{b}}$,
$ \mathcal{L}_{k_{u}}$,
$ \mathcal{L}_{k_{c}}$, and
$ \mathcal{L}_{k_{a}}$ $(k \in \{l, s\})$ denote the L1 loss,
GIOU loss \cite{rezatofighi2019generalized} for bounding box regression,
cross-entropy loss for object classification,
and focal loss \cite{lin2017focal} for interaction prediction,
respectively. These loss functions are realized in the same way as those in \cite{zhang2021mining}. Moreover, $\alpha$, $\beta$, and $\gamma $ are set as 1 for simplicity;
while $\lambda_{b}$, $\lambda_{u}$, $\lambda_{c}$, and $\lambda_{a}$ are set as 2.5, 1, 1, and 1, which are the same values as those in \cite{zhang2021mining}.

\section{Experiments}
\label{sec:Experiments_setting}
\subsection{Datasets and Evaluation Metrics}
\textbf{HICO-DET.} HICO-DET  \cite{chao2018learning} is a large-scale HOI detection dataset that comprises over 150,000 annotated instances. It contains  38,118 and 9,658 images for training and testing, respectively. There are a total of 80 object categories, 117 verb categories, and 600 HOI categories. Those HOI categories with fewer than 10 training samples are referred to as ``rare'' categories and and the remainder  are called as ``non-rare'' categories; specifically, there are a total of 138 rare and 462 non-rare categories. There are two modes of mAP on HICO-DET, namely the Default (DT) mode and the Known-Object (KO) mode. In DT mode, each HOI category is evaluated on all testing images; while in KO mode, one HOI category is only evaluated on images that contain its associated object category.

\textbf{V-COCO.}  V-COCO was  composed based on the MS-COCO database \cite{lin2014microsoft}. The training and validation sets of V-COCO consist of 5,400 images, whereas its testing set includes 4,946 images.  The database covers 80 object categories, 26 interaction categories, and 234 HOI categories. The mean average precision of Scenario 1 role ($mAP_{role}$) \cite{gupta2015visual} is used for evaluation.

\subsection{Implementation Details}

We employ ResNet-50 as the backbone model and adopt the AdamW optimizer. The  experiments are conducted with a batch size of 16 on 8 GPUs. The initial learning rate is set to 1e-4 and then multiplied by 0.1 after 60 epochs; the total number of epochs is 90.
$N_{q}$ is set as 64 and 100 on HICO-DET and V-COCO, respectively; while $D$ is 256.
We initialize the network with the  DETR \cite{carion2020end}  parameters trained on the MS-COCO database \cite{lin2014microsoft}.  
As for the hyper-parameters,
$T_h$, $T_o$, and $T_g$  in SCA are set as 2, 2, and 4,
respectively. 
\begin{table}[t]
  \small
  \setlength{\tabcolsep}{4pt}
  \centering
  \caption{Ablation studies on each component of DIR. For HICO-DET, the DT mode  is adopted for evaluation.}
  \resizebox{0.94\linewidth}{!}{
        \begin{tabular}{cc ccccc cc}
			\hline
			& \multicolumn{5}{c}{Components}  &\multicolumn{2}{c}{mAP}   \\
			Method     &SCA  & IPE &Larger Decoder &Late Fusion  &Early Fusion & HICO-DET   & V-COCO  \\
			\hline
		     Baseline&-&-&-&- &-   &31.44       &61.68 \\
		     \hline
			\multirow{4}{*}{Incremental }
			
			&\checkmark&-&-&-&-     &32.58        &62.78  \\
            &-&\checkmark&-&-&-    &32.70 &63.28  \\
            &-&-&\checkmark&-&- &31.74  &62.11  \\
           &\checkmark&\checkmark&-    &\checkmark &- &32.93      & 63.77    \\
           \hline
            Our Method&\checkmark&\checkmark&- &- &\checkmark&\textbf{33.20}        &\textbf{64.10}     \\
			\hline
	   \end{tabular}}
    \vspace{-0.4cm} 
  \label{tab:tab1}%
\end{table}

\subsection{Ablation Studies}
We perform ablation studies on HICO-DET and V-COCO to demonstrate the effectiveness of each proposed component. We adopt CDN \cite{zhang2021mining} as our baseline and perform all experiments using ResNet-50 as the backbone. Experimental results are presented in Table~\ref{tab:tab1}.

\noindent{\bf{Effectiveness of SCA.}} 
SCA is introduced to encourage the model to extract human appearance, object appearance, and global features in a disentangled way. When SCA is used, the performance of CDN is promoted by 1.04\% and 1.18\% mAP on HICO-DET and V-COCO, respectively.
These results demonstrate the superiority of SCA in improving the performance of one-stage HOI detectors.
We further evaluate the optimal values of $T_h$, $T_o$, and $T_g$ in the supplementary material.

\noindent{\bf{Effectiveness of IPE.}}  IPE is  proposed to efficiently encode the human pose feature for one-stage methods. As illustrated in Table~\ref{tab:tab1}, when equipped with IPE, the performance of CDN is notably improved by 0.97\% and 1.40\% mAP on HICO-DET and V-COCO, respectively. Furthermore, we increase the number of interaction decoder layers in CDN-S by a factor of two, denoted as `Larger Decoder' in Table \ref{tab:tab1}. It is shown the performance of this model only slightly better than that of the baseline model. This result further demonstrates  that the performance promotion achieved by IPE is due to the introduced pose features rather than the increased model complexity.

We here investigate two possible strategies that combine the appearance feature stream and the pose feature stream, namely, Late Fusion and Early Fusion.

\textbf{Late Fusion.} In this strategy, we fuse the appearance stream and pose stream through score addition in line with previous two-stage methods \cite{gupta2019no,zhong2020polysemy}. In more detail, we employ two interaction classifiers with the appearance feature $\textbf{C}^{a}$ and the pose feature $\textbf{C}^{p}$ as input, respectively.  Their classification scores are added to produce the final interaction scores. As shown in Table~\ref{tab:tab1}, this model achieves better performance than the model using SCA or IPE alone. 

\textbf{Early Fusion.} In this strategy, $\textbf{C}^{a}$ and $\textbf{C}^{p}$ are fused through element-wise addition, as shown in  Fig. \ref{Figure:overview}.  We can observe that this setting outperforms the ‘Late Fusion’ setting by 0.27\% and 0.33\% mAP on HICO-DET and V-COCO, respectively. Therefore, we consistently adopt early fusion in the remainder of this paper.

\noindent\textbf{Application to Other One-Stage Models.} Both SCA and IPE are plug-and-play methods that can be readily applied to other DETR-based HOI detection models  \cite{liao2022gen}. The main difference between GEN-VLKT  \cite{liao2022gen}  and CDN \cite{zhang2021mining} is that GEN-VLKT performs human detection and object detection in parallel branches with independent queries. We herein utilize human queries as pose queries of the pose decoder. As shown in Table \ref{tab:hico}, we can see that GEN-VLKT+DIR outperforms GEN-VLKT by 1.04\% mAP in DT mode for the full HOI categories. 

\begin{table}[t]
\centering
\begin{minipage}{0.48\textwidth}
    \centering
\caption{Comparisons with variants of SCA on HICO-DET. FPS stands for Frames Per Second.}

 \resizebox{0.98\linewidth}{!}{
            \begin{tabular}{l| ccc|c }
        		\hline
        			&		Full& Rare&Non-rare & FPS $\uparrow$
\\
        		\hline
        		\hline
        	CDN \cite{zhang2021mining}   &31.44 & 27.39 & 32.64 &66.15\\
        		\hline
          w/o GT boxes  &32.05  &26.85 &33.60 &66.15\\
       Inference w/ $\textbf{M}$ &32.51	&27.86 &33.89 &13.08\\	
          w/o $\textbf{Q}^{s}$ &31.29 &26.26 &32.80  &66.15\\
        	 \hline
          CDN + SCA &   \textbf{32.58}	&\textbf{28.10} &\textbf{33.92} &66.15 \\
        		\hline
    		\end{tabular}
		}
	\label{table:SCA}
    \end{minipage}
     \hfill
    \begin{minipage}{0.48\textwidth}
   \centering
       \caption{Comparisons with variants of IPE on HICO-DET.}
         \resizebox{0.92\linewidth}{!}{
            \begin{tabular}{l|ccc|c}
        		\hline
        			&		Full& Rare&Non-rare & FPS  $\uparrow$\\
        		\hline
        		\hline
        	CDN \cite{zhang2021mining}  &31.44 & 27.39 & 32.64 &66.15\\
        		\hline
        	 w/o $\textbf{M}^{p}$   &32.57  &28.57  &33.76   &65.25\\  
        w/ $\textbf{C}^{p}$ &32.09  &28.17  &32.96 &65.25\\
        w/o $\mathcal{L}_{p}$  &31.88   &28.33 &32.94  &65.25\\
        	 \hline
        	 CDN + IPE & \textbf{32.70}  &\textbf{28.60} &\textbf{33.92} &65.25\\
        		\hline
    		\end{tabular}
		}
		\vspace{-6mm}
	\label{table:IPE}
    \end{minipage}
\end{table}

\subsection{Comparisons with Variants of SCA and IPE}
\subsubsection{Comparisons with Variants of SCA} 
We compare the performance of SCA with its three possible variants, namely,  `w/o GT Boxes', `Inference w/ $\textbf{M}$', and `w/o $\textbf{Q}^{s}$'.
Experimental results are tabulated in Table \ref{table:SCA}. 

First, `w/o GT Boxes' represents that we leverage the predicted bounding boxes by $\textbf{Q}^{s}$ to generate the head masks for the interaction  decoder layers during training. We can observe that without GT boxes, the performance of SCA is dropped by 0.53\%, 1.25\%, and 0.32\% mAP in DT mode for the full, rare, and non-rare HOI categories, respectively. This experimental result demonstrates the effectiveness of using GT bounding boxes for implementing SCA. 

Second, `Inference w/ $\textbf{M}$' means that we use the predicted bounding boxes by $\textbf{Q}^{l}$ to generate the head masks for the interaction  decoder layers during inference.  As shown in Table  \ref{table:SCA}, the performance of this variant is comparable to that of SCA. However, it is important to note that the inference speed of this variant significantly decreases due to the additional computational cost involved. This result further demonstrates the superiority of our proposed SCA mechanism.

Finally, `w/o $\textbf{Q}^{s}$' means that we directly apply the shunted cross-attention to the learnable queries $\textbf{Q}^{l}$. This indicates that there is only one set of HOI queries $\textbf{Q}^{l}$ during both training and inference. We utilize the predicted bounding boxes by $\textbf{Q}^{l}$ to generate the shunted head mask.
It is shown that the performance of this variant is even lower than CDN by 0.15\% and 1.13\%  mAP in DT mode for the full and rare  HOI categories, respectively. As analyzed in Section \ref{sec:SCA}, the predicted bounding boxes by  $\textbf{Q}^{l}$ may be in poor-quality at the beginning of the training phase and therefore bring in interference to the interaction representation learning.


\subsubsection{Comparisons with Variants of IPE} 
We compare the performance of IPE with some possible variants. Experimental results are tabulated in Table \ref{table:IPE}. 

First, `w/o $\textbf{M}^{p}$'  refers to removing the keypoint mask $\textbf{M}^{p}$ in Eq.(\ref{eq:pose_loss}). It can be seen that the performance of this variant is lower than our IPE by 0.23\% mAP in DT mode for the full HOI categories. As discovered in previous works \cite{li2020pastanet}, each interaction category is usually related to only part of the keypoints. This experimental result provides direct evidence that IPA is effective at capturing interaction-aware keypoints.

\begin{table}[t]
\centering
\begin{minipage}{0.6\textwidth}
    \centering
\caption{Performance comparisons on HICO-DET.}

\resizebox{0.98\textwidth}{!}{
\begin{tabular}{c|c|c|ccc  }
\hline
&    & & \multicolumn{3}{c}{DT Mode} \\
& Methods & Backbone & Full & Rare & Non-Rare  \\
\hline
\hline
\multirow{5}*{\rotatebox{90}{Two-Stage}}
&PMFNet \cite{wan2019pose} &ResNet-50-FPN & 17.46 &15.65 &18.00 \\
&TIN \cite{li2019transferable}& ResNet-50    &17.22 &13.51 &18.32 \\
& PD-Net \cite{zhong2020polysemy}  & ResNet-152  &20.81 &15.90 &22.28\\
& DJ-RN \cite{li2020detailed} & ResNet-50  &21.34 &18.53 &22.18 \\
& SCG\cite{scg2021}  & ResNet-50  &31.33 &24.72 &33.31 \\
\hline
\hline
\multirow{12}*{\rotatebox{90}{One-Stage}}
&PPDM \cite{liao2020ppdm}& Hourglass-104   & 21.73 &13.78 &24.10 \\

&HOTR \cite{kim2021hotr}  & ResNet-50  &23.46 &16.21 &25.62 \\
&HOI-Trans \cite{zou2021end}   & ResNet-50 & 23.46 & 16.91 & 25.41 \\
& GGNet \cite{zhong2021glance}    & Hourglass-104   &23.47 &16.48 &25.60\\
 &AS-Net \cite{kim2020uniondet}   & ResNet-50 &28.87  & 24.25  & 30.25 \\
&QPIC  \cite{tamura2021qpic}  & ResNet-50   &29.07 &21.85 &31.23  \\
&CDN-S \cite{zhang2021mining}     & ResNet-50  &31.44 & 27.39 & 32.64 \\
&  HQM  \cite{zhong2022towards} & ResNet-50 & 32.47 &28.15 & 33.76  \\
&  IF  \cite{liu2022interactiveness} & ResNet-50 & 33.51 &30.30 &34.46  \\
&GEN-VLKT$_s$ \cite{liao2022gen}& ResNet-50  & 33.75 & 29.25 & 35.10   \\
&CDN-S \cite{kim2021hotr} + \textbf{Ours}     & ResNet-50  & \textbf{33.20} & \textbf{29.96} &  \textbf{34.17}  \\
&GEN-VLKT$_s$ \cite{liao2022gen} + \textbf{Ours}  & ResNet-50  &\textbf{34.79} &\textbf{31.80} &\textbf{35.68} \\
\hline
\end{tabular}}
\label{tab:hico}
    \end{minipage}
     \hfill
    \vspace{-2mm}
    \begin{minipage}{0.38\textwidth}
    \centering
       \caption{Performance comparisons on V-COCO.}
        \resizebox{0.98\textwidth}{!}{
        \begin{tabular}{c|c|c|c}
        \hline
        &Methods    & Backbone   & $AP_{role}$ \\
        \hline 
        \hline
        \multirow{7}*{\rotatebox{90}{Two-Stage}}
        &InteractNet \cite{Gkioxari2017Detecting} &ResNet-50-FPN  &40.0 \\
        & VCL \cite{hou2020visual}  & ResNet-101  &48.3 \\
        &DRG \cite{Gao-ECCV-DRG}  &ResNet-50-FPN &51.0 \\ 
        & VSGNet \cite{ulutan2020vsgnet} &Res152 & 51.7\\
        &PMFNet \cite{wan2019pose} &ResNet-50-FPN & 52.0 \\
        &PD-Net \cite{zhong2020polysemy} &ResNet-152 &52.6\\
        &ACP \cite{kim2020detecting} &ResNet-152 &52.9\\ 
        \hline
        \hline
        \multirow{10}*{\rotatebox{90}{One-Stage}}
        &HOI-Trans \cite{zou2021end} &ResNet-101  &52.9 \\
        &AS-Net \cite{chen2021reformulating} &ResNet-50 &53.9\\
         & GGNet \cite{zhong2021glance}    &Hourglass-104 & 54.7 \\ 
        &HOTR \cite{kim2021hotr} &ResNet-50  &55.2 \\
        &QPIC  \cite{tamura2021qpic} & ResNet-50 &58.8\\
         &CDN-S  \cite{zhang2021mining} & ResNet-50 &61.7\\
         &GEN-VLKT$_s$ \cite{liao2022gen}& ResNet-50 &62.4\\
          &DOQ \cite{qu2022distillation}  & ResNet-50 &63.5 \\
          &HQM \cite{zhong2022towards}  & ResNet-50 &63.6 \\
           &CDN-S \cite{zhang2021mining}  + \textbf{Ours}  & ResNet-50 &\textbf{64.1} \\
           &GEN-VLKT$_s$ \cite{liao2022gen} + \textbf{Ours}  & ResNet-50  &\textbf{64.5} \\
        \hline
        \end{tabular}}
        \label{VCOCO}
    \end{minipage}
\end{table}

Second, `w/ $\textbf{C}^{p}$'  represents that we sent the pose feature $\textbf{C}^{p}$ to IPA in Eq.(\ref{eq:IPA}) for the prediction of  keypoint weights rather than using $\textbf{Q}^{int}$. Compared with using $\textbf{Q}^{int}$, the performance of this variant is lower than our IPA by 0.61\%, 0.43\%, and 0.96\% mAP in DT mode for the full, rare, and non-rare HOI categories, respectively. 
The main reason may be that $\textbf{Q}^{int}$ contains more interaction cues than that of $\textbf{C}^{p}$, which allows for better inference of interaction-aware keypoints. 

Finally, `w/o $\mathcal{L}_{p}$' means we remove the pose loss $\mathcal{L}_{p}$ from IPE. It is shown that the performance of this variant is lower than our IPE by 0.82\%, 0.27\%, and 0.98\% mAP in DT mode for the full, rare, and non-rare HOI categories, respectively. This result further demonstrates that the performance improvement achieved by IPE is attributed to the introduced pose features rather than the increased complexity of the model.

\subsection{Comparisons with State-of-the-art Methods}
\noindent{\bf{Comparisons on HICO-DET}}. As shown in Table \ref{tab:hico}, our method outperforms state-of-the-art methods by considerable margins. Impressively, CDN-S+DIR outperforms CDN-S \cite{zhang2021mining} by 1.76\%, 2.57\%, and 1.53\% mAP on the full, rare, and non-rare HOI categories of the DT mode, respectively. When our method is applied to GEN-VLKT \cite{liao2022gen}, with ResNet-50 as the backbone, GEN-VLKT+DIR achieves a promising performance of 34.79\% mAP. These experiments justify the effectiveness of DIR in promoting the power of interaction representations for one-stage methods. Comparisons under the KO mode are presented in the supplementary material.

\noindent{\bf{Comparisons on V-COCO}}. Comparisons on V-COCO are tabulated in Table \ref{VCOCO}.  It can be observed that our method still outperforms state-of-the-art approaches, achieving 64.5\% in terms of $AP_{role}$. These experiments demonstrate that DIR consistently improves the quality of interaction representations for the one-stage models on both the HICO-DET and VCOCO databases.

\begin{figure*}[t]
    \centering
\includegraphics[width=1.0\textwidth]{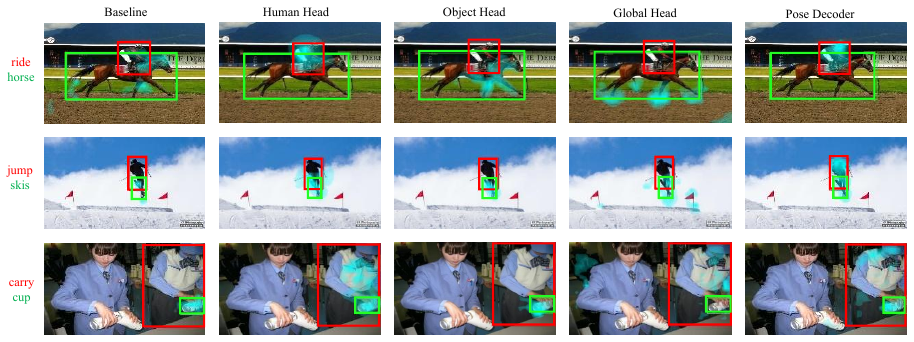}
\caption{Visualization of the cross-attention maps in one decoder layer on HICO-DET. Images in the first column  represents the attention maps for the interaction decoder of CDN-S \cite{zhang2021mining}; while the remaining four columns represent cross-attention maps for human, object, global feature heads, and pose decoder of CDN-S+DIR, respectively. Best viewed in color.}
    \label{Figure:Visualization}
    \vspace{-2mm}
\end{figure*}

\subsection{Qualitative Visualization Results}
As presented in Fig. \ref{Figure:Visualization}, we visualize some HOI detection results and the cross-attention maps from CDN-S and CDN-S+DIR. First, as the figure shows, our DIR can effectively disentangle and learn human appearance, object appearance, and global feature by using the SCA training mechanism. Moreover, as shown in the figure, it is evident that IPE can assist the model to attend to important human body keypoints. Finally, by integrating a variety of interpretable cues into a disentangled interaction feature, the one-stage model’s prediction accuracy is enhanced during inference. More qualitative comparison results are provided in the supplementary material.

\section{Conclusion}
\label{others}
In this paper, we propose a novel approach that enables one-stage methods to extract disentangled interaction representations, namely, Disentangled Interaction Representations (DIR). Our DIR approach mimics the disentangled interaction feature learning of two-stage methods. To facilitate the learning of human and object appearance features in a disentangled manner, we propose a novel training mechanism named Shunted Cross-Attention (SCA). SCA can be abandoned during inference, meaning that no additional cost is introduced. To integrate the pose feature, we introduce a novel Interaction-aware Pose Estimation (IPE) task. IPE efficiently extracts the pose features that are relevant to the interaction between the interested human-object pair. DIR can be readily applied to existing one-stage HOI detection methods. Extensive experiments are conducted to demonstrate the effectiveness of each proposed component. Finally, we consistently achieve state-of-the-art performance on two benchmarks: HICO-DET and V-COCO.

\section*{Broader Impacts}
Similar to many other AI technologies, our proposed DIR has inherent potential for both positive and negative impacts. While the technology itself is not harmful, it is crucial to acknowledge that any tools or systems can be misused or repurposed for malicious activities. As responsible developers and researchers, we emphasize the importance of ethical and conscientious adoption of our technique.

\clearpage

{
\small
\bibliographystyle{plain}
 
}

\clearpage

\section*{Supplementary Material}

\begin{appendix}
This supplementary material includes five sections. 
Section \ref{sec:1} illustrates the structure of DIR in the inference stage. Section \ref{sec:2} provides a comprehensive description of the detailed designs for SCA.
Section \ref{Ablation} conducts ablation study on the value of some hyper-parameters in our approach.  Section \ref{sec:hico} tabulates the complete comparison results on HICO-DET. Section \ref{vis} provides qualitative comparison results for CDN-S and CDN-S+DIR.

\section{Overview of DIR During Inference}
\label{sec:1}
Figure \ref{Figure:overview} illustrates the structure of DIR in the inference stage.
During inference, only $\textbf{Q}^{l}$ are sent to the interaction decoder and pose decoder to produce $\textbf{C}^{a}$ and $\textbf{C}^{p}$, respectively; therefore, the SCA component and IPA are removed.
\begin{figure*}[!h]
	\centering
		\includegraphics[width=1.0\textwidth]{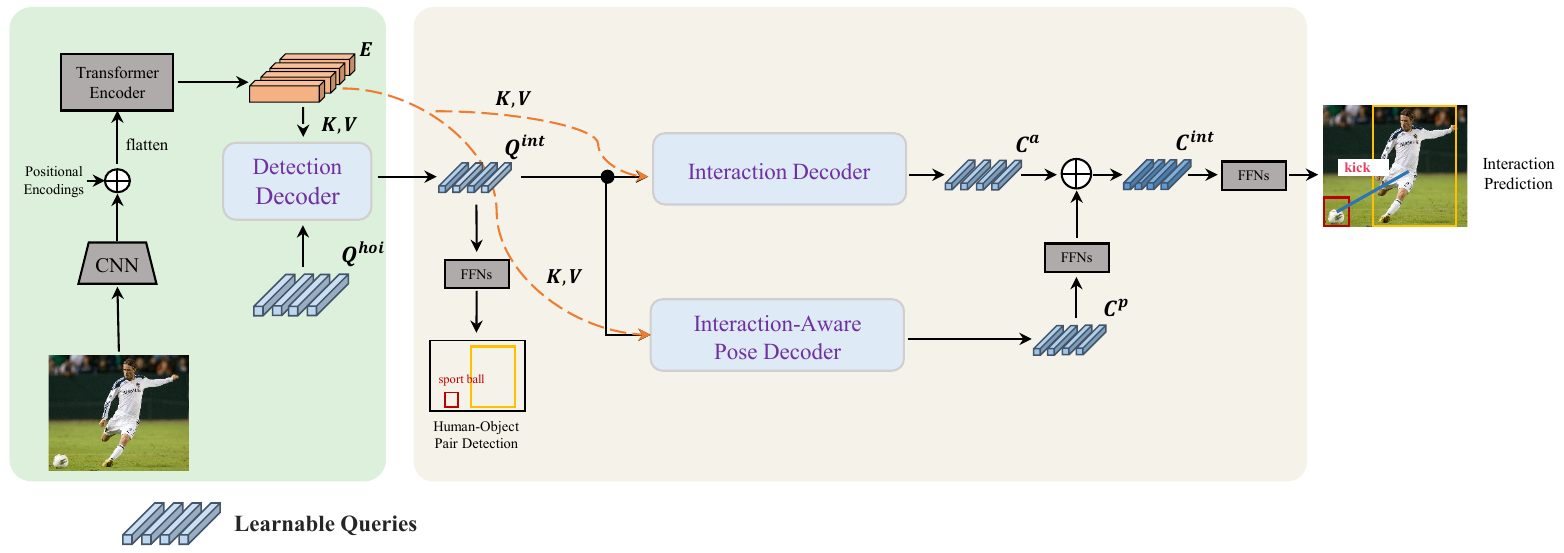}
	\caption{Overview of DIR in the inference stage. $\textbf{Q}^{hoi}$ are the queries for the detection decoder, which is responsible for interactive human-object pair detection. $\textbf{Q}^{hoi}$ only contain the learnable queries $\textbf{Q}^{l}$. $\textbf{Q}^{int}$ are the queries for both the interaction and pose decoders, which output the appearance feature $\textbf{C}^{a}$ and the pose feature $\textbf{C}^{p}$. The final interaction feature is obtained by fusing $\textbf{C}^{p}$ and $\textbf{C}^{a}$ via element-wise addition. Best viewed in color.}
	\label{Figure:overview}
	\vspace{-3mm}
\end{figure*}

\section{Model Structure of Shunted Cross Attention}
\label{sec:2}
\begin{figure*}[t]
    \centering
    \includegraphics[width=0.88\textwidth]{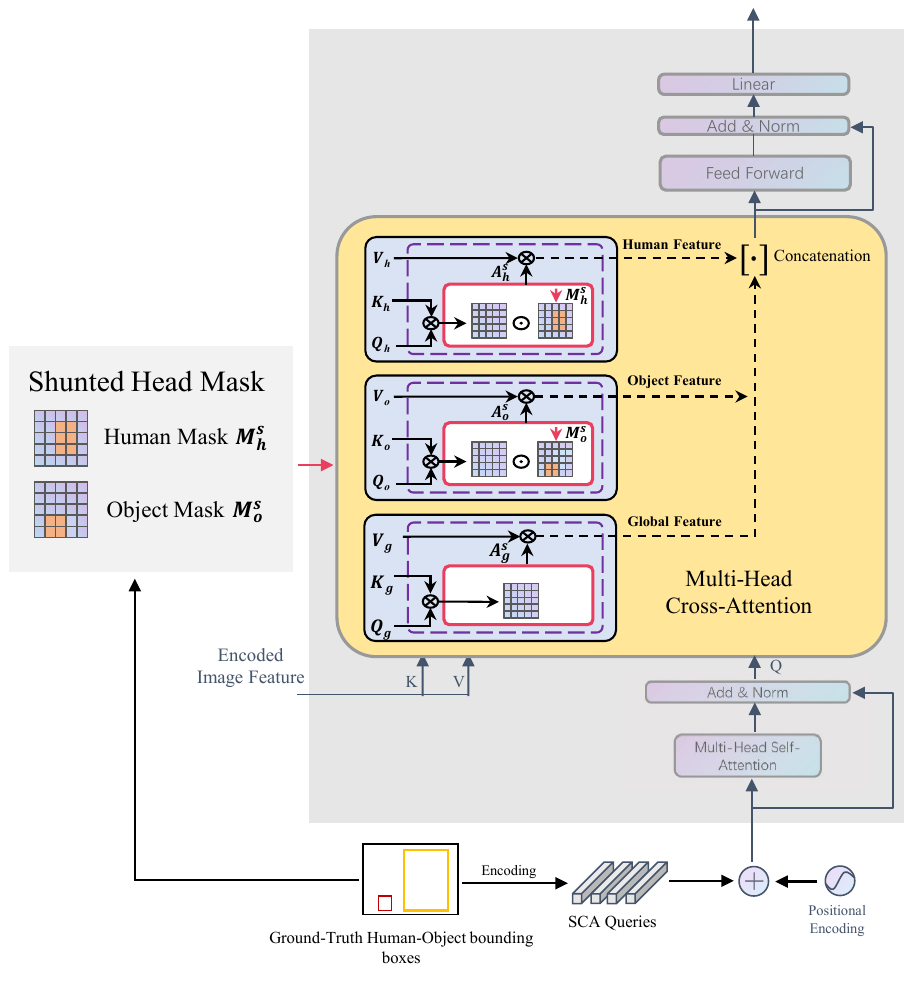}
   \caption{Model Structure of Shunted Cross Attention. $\odot$ and $\oplus$ denote the element-wise multiplication and addition operations,  respectively; while $\otimes$ represents matrix multiplication. Best viewed in color.}
    \label{Figure:SCA}
    \vspace{-1mm}
\end{figure*}

Figure \ref{Figure:SCA} provides the detailed designs for the model structure of SCA.  During training, the coordinates of the ground-truth human-object pairs are used to generate both the SCA queries and the shunted head masks. The shunted head masks are imposed to different cross-attention heads to learn human appearance feature, object appearance feature and global feature, respectively.

\section{Ablation Study on Hyper-parameters}
\label{Ablation}

\subsection{Ablation Study on Hyper-parameters of SCA}
\begin{table}[h]
\centering
\begin{minipage}{0.48\textwidth}
    \centering
\caption{Ablation Study on the Value of $T_h$, $T_o$, and $T_g$ in SCA.}
\resizebox{0.92\linewidth}{!}{
            \begin{tabular}{ccc| ccc}
        		\hline
        			\# $T_h$  &\# $T_o$  &\# $T_g$ 		&Full& Rare&Non-rare\\
        		\hline
        		\hline
        		  	1 &1 &6  &32.21 &27.98 &33.57  \\
        	 	 2	&2  &4   &\textbf{32.58}	&\textbf{28.10} &\textbf{33.92} \\
                 3    &3  &2 &32.28 & 28.10 & 33.30 \\
        		\hline
    		\end{tabular}
		}
	\label{table:SCA}
    \end{minipage}
     \hfill
    \begin{minipage}{0.48\textwidth}
   \centering
       \caption{Ablation Study on the Value of  $\gamma$ and $\mathcal{L}_{p}$ for IPE.}
         \resizebox{0.7\linewidth}{!}{
            \begin{tabular}{cc| ccc}
        		\hline
        			  $\gamma$	&$\mathcal{L}_{p}$	&Full& Rare&Non-rare\\
        		\hline
        		\hline
        		0.5  &L1  &31.64  &27.20 &32.97  \\
        	 	 	1.0 &L1 &\textbf{32.70}  &\textbf{28.60} &\textbf{33.92} \\
              1.5 &L1 &32.35  &28.64  &33.45 \\
        		\hline
          1.0 &L2 &32.33  &27.82 &33.68 \\
          \hline
    		\end{tabular}
		}
		\vspace{-2mm}
	\label{table:IPE}
    \end{minipage}
\end{table}

Experiments are conducted on the HICO-DET database and the results are tabulated in Table \ref{table:SCA}.
It is shown that SCA achieves the best performance when $T_h$, $T_o$, and $T_g$ is set as 2, 2, and 4, respectively.

\subsection{Ablation Study on Hyper-parameters of IPE}
Experiments are conducted on the HICO-DET database and the results are tabulated in Table \ref{table:IPE}.
We observe that IPE achieves the best performance when  $\gamma$ and  is set as 1.0 and $\mathcal{L}_{p}$ is implemented using L1 loss, respectively.

\section{Performance Comparisons on HICO-DET}
\label{sec:hico}
We here present the complete comparisons between our method and state-of-the-arts on HICO-DET in Table \ref{tab:hico}.

\begin{table*}[t]
\centering
\caption{Performance comparisons on HICO-DET.}
 \vspace{-3mm}
\resizebox{0.92\textwidth}{!}{
\begin{tabular}{c|c|c|c|ccc|ccc  }
\hline
                &      &        & &\multicolumn{3}{c|}{DT Mode}  &\multicolumn{3}{c}{KO Mode}\\
& Methods   & Backbone &Detector & Full & Rare & Non-Rare  & Full & Rare & Non-Rare \\
\hline
\hline
\multirow{6}*{\rotatebox{90}{Two-Stage}}
&TIN \cite{li2019transferable}& ResNet-50  &HICO-DET   &17.22 &13.51 &18.32     & 19.38 &15.38 &20.57 \\
&UnionDet \cite{kim2020uniondet}   & ResNet-50-FPN  &HICO-DET &17.58  & 11.72  & 19.33  & 19.76  & 14.68  & 21.27 \\
& PD-Net \cite{zhong2020polysemy}   & ResNet-152  &COCO &20.81 &15.90 &22.28 &24.78 &18.88 &26.54\\
& DJ-RN \cite{li2020detailed}  & ResNet-50  &COCO &21.34 &18.53 &22.18 &23.69 &20.64 &24.60 \\
& VCL \cite{hou2020visual}   & ResNet-101  &HICO-DET &23.63 &17.21 &25.55 &25.98 &19.12 &28.03 \\
& SCG \cite{scg2021}  & ResNet-50 &HICO-DET &31.33 &24.72 &33.31 & -   & -   & - \\
\hline
\hline
\multirow{10}*{\rotatebox{90}{One-Stage}}
&PPDM \cite{liao2020ppdm}   & Hourglass-104  &HICO-DET & 21.73 &13.78 &24.10  & 24.58   & 16.65   & 26.84\\
& GGNet \cite{zhong2021glance}        & Hourglass-104  &HICO-DET &23.47 &16.48 &25.60 &27.36 &20.23 & 29.48 \\
&HOTR \cite{kim2021hotr}   & ResNet-50  &COCO &23.46 &16.21 &25.62     & -   & -   & - \\
&HOI-Trans \cite{zou2021end}     & ResNet-50 &HICO-DET & 23.46 & 16.91 & 25.41 & 26.15 & 19.24 & 28.22 \\
 &AS-Net \cite{kim2020uniondet}    & ResNet-50 &HICO-DET &28.87  & 24.25  & 30.25  & 31.74  & 27.07  & 33.14 \\
&QPIC  \cite{tamura2021qpic}    & ResNet-50  &HICO-DET &29.07 &21.85 &31.23  & 31.68   & 24.14   & 33.93 \\
&CDN-S \cite{zhang2021mining}     & ResNet-50 &HICO-DET &31.44 & 27.39 & 32.64 &34.09 &29.63 &35.42 \\
& HQM \cite{zhong2022towards}     & ResNet-50  &HICO-DET & 32.47 & 28.15 &33.76  &35.17 &30.73 &36.50 \\
&GEN-VLKT$_s$ \cite{liao2022gen}& ResNet-50  &HICO-DET & 33.75 & 29.25 & 35.10 &36.78  &32.75 &37.99     \\
&CDN-S \cite{zhang2021mining} + \textbf{Ours}     & ResNet-50 &HICO-DET  & \textbf{33.20} & \textbf{29.96} &  \textbf{34.17} & \textbf{36.10}  & \textbf{32.87}  & \textbf{37.61} \\
&GEN-VLKT \cite{liao2022gen} + \textbf{Ours}   &ResNet-50  &HICO-DET &\textbf{34.79} &\textbf{31.80} &\textbf{35.68}  & \textbf{38.24}  & \textbf{35.51}  & \textbf{39.06} \\

\hline
\end{tabular}}
\vspace{-3mm}
\label{tab:hico}
\end{table*}

\begin{figure*}[h]
    \centering
    \includegraphics[width=1.0\textwidth]{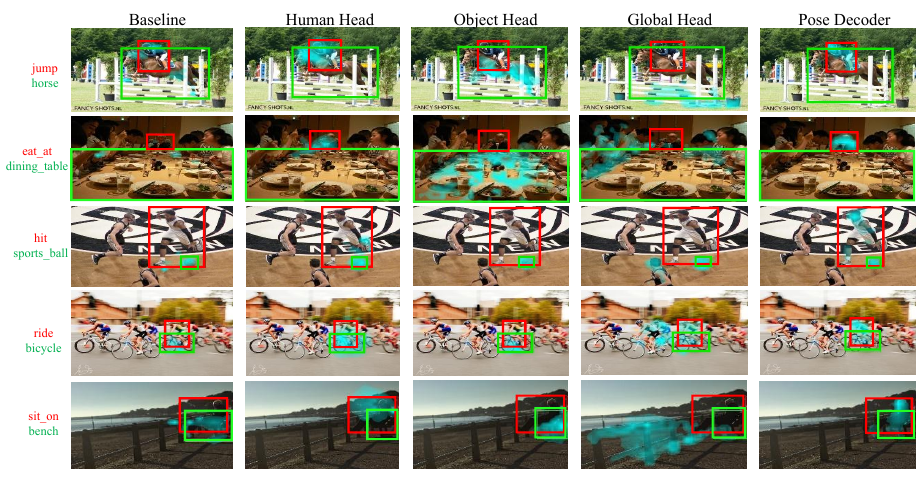}
   \caption{Visualization of the cross-attention maps in one decoder layer on HICO-DET. Images in the first column  represents the attention maps for the interaction decoder of CDN-S \cite{zhang2021mining}; while the remaining four columns represent cross-attention maps for human, object, global feature heads, and pose decoder of CDN-S+DIR, respectively. Best viewed in color.}
    \label{Figure:Vis_supp}
    \vspace{-1mm}
\end{figure*}

\section{Qualitative Visualization Results}
\label{vis}
Fig. \ref{Figure:Vis_supp} provides more qualitative comparisons between
CDN-S \cite{tamura2021qpic} and CDN-S+DIR  in terms of cross-attention maps and HOI detection results on HICO-DET. It is evident that the DIR enables CDN-S to effectively disentangle and integrate diverse interpretable cues, leading to improved accuracy in HOI detection.

\end{appendix}

\end{document}